\documentclass[twoside]{article}
 
\usepackage[accepted]{aistats2015}

\usepackage{graphicx}
\usepackage[sort, round]{natbib}
\usepackage{algorithm}
\usepackage{algpseudocode}
\usepackage{listings}
\usepackage[draft]{pgf}
\usepackage{caption, subcaption}
\usepackage{amsmath}
\usepackage{amssymb}
\usepackage{xfrac}
\usepackage{fancyvrb}
\usepackage{color}
\usepackage{lstlinebgrd}
\usepackage{pgffor}
\usepackage{dsfont}
\usepackage{todonotes}
\usepackage{microtype}
\usepackage{hyperref}

\graphicspath{{./figures/}}

\definecolor{lightgray}{rgb}{0.9,0.9,0.9}
\definecolor{darkgray}{rgb}{0.4,0.4,0.4}
\definecolor{darkolivegreen}{rgb}{0.33, 0.42, 0.18}
\definecolor{darklava}{rgb}{0.28, 0.24, 0.2}
\definecolor{lightblue}{rgb}{0.68, 0.85, 0.9}
\definecolor{smalt(darkpowderblue)}{rgb}{0.0, 0.2, 0.6}
\definecolor{richelectricblue}{rgb}{0.03, 0.57, 0.82}
\definecolor{ultramarineblue}{rgb}{0.25, 0.4, 0.96}
\definecolor{bleudefrance}{rgb}{0.19, 0.55, 0.91}
\definecolor{dimgray}{rgb}{0.75, 0.75, 0.75}

\algrenewcommand\algorithmicindent{1.0em}%
\newcommand{\ind}{\hspace{\algorithmicindent}}

\newcommand{\address}{\mathbf{address}}
\newcommand{\level}{\mathbf{level}}

\usepackage{letltxmacro}
\newcommand*{\SavedLstInline}{}
\LetLtxMacro\SavedLstInline\lstinline
\DeclareRobustCommand*{\lstinline}{%
  \ifmmode
    \let\SavedBGroup\bgroup
    \def\bgroup{%
      \let\bgroup\SavedBGroup
      \hbox\bgroup
    }%
  \fi
  \SavedLstInline
}

\lstdefinelanguage{JavaScript}{
  keywords={typeof, new, true, false, catch, function, return, null, catch, switch, var, if, in, while, do, else, case, break},
  keywordstyle=\bfseries,
  ndkeywords={class, export, boolean, throw, implements, import, this},
  ndkeywordstyle=\bfseries,
  identifierstyle=\color{black},
  sensitive=false,
  comment=[l]{//},
  morecomment=[s]{/*}{*/},
  commentstyle=\color{darkolivegreen}\ttfamily,
  stringstyle=\color{red}\ttfamily,
  morestring=[b]',
  morestring=[b]"
}

\lstdefinestyle{hilight}{
  moredelim=**[is][\color{black}]{@}{@},
}

\begin{document}

\twocolumn[

\aistatstitle{Coarse-to-Fine Sequential Monte Carlo for Probabilistic Programs}

\aistatsauthor{ Andreas Stuhlm\"uller \ \ Robert X.D. Hawkins \ \ N. Siddharth \ \ Noah D. Goodman }

\aistatsaddress{ Department of Psychology \\ Stanford University \\ \{astu, rxdh, nsid, ngoodman\}@stanford.edu} ]

\begin{abstract}
  Many practical techniques for probabilistic inference require a sequence of distributions that interpolate between a tractable distribution and an intractable distribution of interest.
  Usually, the sequences used are simple, e.g., based on geometric averages between distributions.
  When models are expressed as probabilistic programs, the models themselves are highly structured objects that can be used to derive annealing sequences that are more sensitive to domain structure.
  We propose an algorithm for transforming probabilistic programs to {\em coarse-to-fine programs} which have the same marginal distribution as the original programs, but generate the data at increasing levels of detail, from coarse to fine.
  We apply this algorithm to an Ising model, its depth-from-disparity variation, and a factorial hidden Markov model.
  We show preliminary evidence that the use of coarse-to-fine models can make existing generic inference algorithms more efficient.
\end{abstract}

\section{INTRODUCTION}

\begin{figure}[t]
  \centering
  \includegraphics[scale=.4]{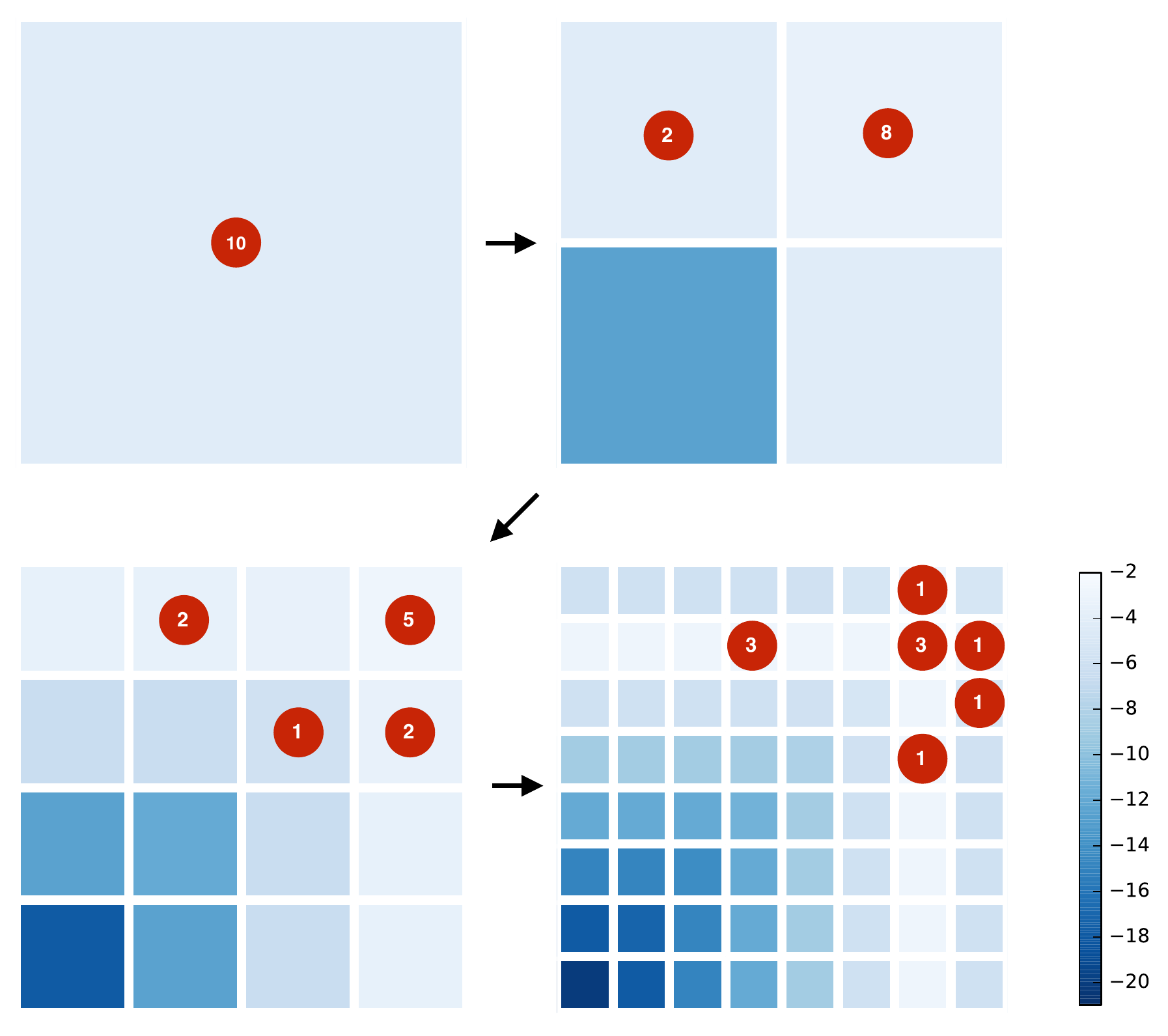}
  \caption{Incremental coarsening reduces surprise in SMC. Particles (red) are directed towards high-probability regions (light) step by step as we refine the state space from coarse to fine. The numbers indicate how many particles are associated with a particular state.}
  \label{fig:ctf-smc-illustration}
\end{figure}%

Imagine watching a tennis tournament.
Your visual system makes fast and accurate inferences about the depth-field (how far away are different patches?), the objects (is that a ball or racket?), their trajectories, and many other properties of the scene.
A powerful intuition is that such feats of inference are enabled by \emph{coarse-to-fine} reasoning: first getting a rough sense of where the action is in the scene, about how far away it is, and so on; later refining this impression to pick out details.
The appeal of coarse-to-fine reasoning is manifold.
First, there is introspection: When faced with a complex reasoning task, it often helps to take a step back, try to understand the big picture, and then focus on what seems most promising. The big picture tends to have fewer moving parts, and its parts tend to be easier to understand.
Neuroscience provides another angle: for instance, face processing in the high-level visual cortex plausibly follows coarse-to-fine principles \citep{goffaux2010coarse}, and stereoscopic depth perception similarly proceeds from large to small spatial scales \citep{menz2003stereoscopic}.
Finally, there is a rich set of existing applications of coarse-to-fine techniques for specific applications in a diverse set of areas including physical chemistry \citep{Lyman2006}, speech processing \citep{Tang2006}, PCFG parsing \citep{Charniak:2006ut}, and machine translation \citep{Petrov2008}.
Despite the success and appeal of coarse-to-fine ideas, they have been difficult to apply in general settings.
Here we propose a system for deriving coarse-to-fine inference from any model written as a probabilistic program.
We do this by leveraging program structure to transform the initial program into a multi-level coarse-to-fine program that can be used with existing inference algorithms.

Probabilistic programming languages provide a universal and high-level representation for probabilistic models, separating the burdens of modeling from those of inference.
Yet the difficulty of inference can grow quickly as the state space (number of program executions) grows large.
A widely-used technique for inference in large state spaces is Sequential Monte Carlo (SMC), a class of algorithms based on constructing a sequence of distributions, beginning with an easy-to-sample distribution and ending with the distribution of interest, with each distribution serving as an importance sampler for the next.
The success of SMC rests on the quality of the approximating sequence.
We present a generic method for deriving coarse-to-fine sequences of approximating distributions from a probabilistic program.

Our approach can be seen as building a hierarchical model from an initial model, where each stage of the hierarchy resolves more details of the state space than the one before. We additionally augment each level of the hierarchy with a coarse approximation to the evidence (implemented via \emph{heuristic factors}, see Section \ref{sec:heuristic-factors}), in order to specify a useful conditional distribution at each level.
In practice we create the hierarchical model and heuristic factors at once by specifying how to ``lift'' each element of the program---elementary distributions, factors, primitive functions, and constants---to the coarser levels.
The resulting model supports coarse-to-fine inference by SMC, where the nth distribution in the sequence is simply the state space of the n-coarsest levels; this is correct inference for the original model because, by construction, the marginal distribution over the finest level is the original distribution.

Model transformations let us directly use existing sequential inference algorithms to perform coarse-to-fine inference, rather than proposing a new inference algorithm \emph{per se}.
This is in contrast to essentially all prior work on coarse-to-fine inference, including \citet{kiddon2011coarse} and \citet{steinhardt2014filtering}.
One benefit of this modular approach is that advances in SMC algorithms immediately yield improvements to coarse-to-fine inference.
Another benefit is the conceptual clarity that comes from an explicit representation of the coarse-to-fine model.

In the following, we first review probabilistic programs and Sequential Monte Carlo. We then describe our coarse-to-fine program transform and how it lifts random variables, primitive functions, and factors to operate on multiple levels of abstraction. We apply this transform to two models in the domain of tracking partially observable objects over time given visual information, a depth-from-disparity model and a factorial hidden Markov model, and show preliminary evidence that it may help reduce inference time in these domains. Finally, we discuss the current limitations of this framework, the circumstances where our approach to coarse-to-fine inference is a good fit, and outline research questions raised by this new approach.

\section{BACKGROUND}

\subsection{PROBABILISTIC PROGRAMS}

\begin{figure*}[t]
  \centering
  \begin{subfigure}[b]{.45\textwidth}
    \centering
    \begin{lstlisting}
var noisyObserve = function(obs){
  var score = -3*distance(obs, 7)
  factor(score)
}

var model = function(){
  var x = sample(uniformERP)
  var y = sample(uniformERP)
  var observation = flip(.5) ? x : y
  noisyObserve(observation)
  return [x, y]
}
\end{lstlisting}
    \caption{A probabilistic program}
    \label{fig:probabilistic-program}
  \end{subfigure}
  \hskip 1em
  \begin{subfigure}[b]{.45\textwidth}
    \centering
    \begin{lstlisting}[
      basicstyle=\footnotesize\ttfamily\color{dimgray},
      identifierstyle=\color{dimgray},
      style=hilight]
var noisyObserve = function(obs){
  var score = -3*distance(obs, 7)
  factor(score)
}

var model = function(){
  var x = sample(uniformERP)
  @var heuristicScore = -distance(x, 7)@
  @factor(heuristicScore)@
  var y = sample(uniformERP)
  var observation = flip(.5) ? x : y
  noisyObserve(observation)
  @factor(-heuristicScore)@
  return [x, y]
}
\end{lstlisting}
    \caption{Rewritten using heuristic factors}
    \label{fig:heuristic-factors}
  \end{subfigure}
  \caption{Two probabilistic programs with the same marginal distribution (shown in the final panel in Figure \ref{fig:ctf-smc-illustration}).}
\end{figure*}

Probabilistic programs are models expressed in Turing-complete languages that supply primitives for random sampling and probabilistic inference \citep[e.g.,][]{koller.d:1997, pfeffer200714, Goodman:2008uq}.
Many existing probabilistic models have been expressed concisely as probabilistic programs.
A distinguishing feature of probabilistic programming as a machine learning technique is that it separates inference techniques from modeling assumptions.
Thus, any advances in algorithms provide benefits for a wide range of applications at once.
While we demonstrate our technique for a small set of models chosen for their pedagogical value, we emphasize that the technique can be applied to a much wider range of models without modification.

We express probabilistic programs in {\em WebPPL} \citep{dippl}, a small probabilistic language embedded in Javascript. This language is universal, and feature-rich, so we expect the techniques to generalize straightforwardly to other languages.
In this language, all random choices are marked by \lstinline{sample}; the argument to \lstinline{sample} is a distribution object (also called {\em Elementary Random Primitive}, or {\em ERP}), its return value a sample from this distribution.
Calls to functions such as \lstinline{flip(0.5)} are shorthand for \mbox{\lstinline{sample(bernoulliERP, [0.5])}}.

To enable probabilistic conditioning, the language supports \lstinline{factor} statements.
The argument to \lstinline{factor} is a score: a number that is added to the log-probability of a program execution, thus increasing or decreasing its relative posterior probability. This includes hard conditioning on evidence as a special case (scores $0$ and $-\infty$).
Finally, the language supports inference primitives such as \lstinline{ParticleFilter} and \lstinline{MH} (Metropolis-Hastings).
Each of these takes as an argument a thunk, that is, a stochastic function that itself takes no arguments. And each of these computes or estimates the distribution on return values of this thunk (its {\em marginal distribution}), taking into account the re-weighting induced by factor statements.

Figure \ref{fig:probabilistic-program} shows a program that implements a simplified one-step version of multiple object tracking: the noisily observed value $7$ could have been produced by either $x$ or $y$, each of which is uniformly chosen from $\{1,2,\dots,8\}$. The final panel in Figure \ref{fig:ctf-smc-illustration} shows the marginal distribution on $[x, y]$ for this program.

In probabilistic programs, the same syntactic variable can be used multiple times. The prototypical example is the geometric distribution:
\begin{lstlisting}
var geometric = function() {
  return flip(0.1) ? 0 : 1 + geometric()
}
\end{lstlisting}
The call to \lstinline{flip(0.1)} may occur an unbounded number of times. For many purposes, it is necessary to distinguish and refer to these different calls. In the context of MCMC, \citet{wingate2011lightweight} introduced a suitable naming scheme based on stack addresses. The address of a random choice is a list of syntactic locations, one for each function on the function call stack at the time when the random variable was sampled. We will build on this scheme to associate corresponding random choices on different levels of coarsening with each other, and use $\address$ in the following to refer to the current stack address.

\subsection{SEQUENTIAL MONTE CARLO}

Suppose our target distribution is $X$ with probability mass function $p$.
Importance sampling generates samples from an approximating distribution $Y$ (with probability mass function $q$) and re-weights the samples to account for the difference between true and approximating $w(x) = p(x)/q(x)$.
To compute estimates of $\psi = \mathbb{E}_{x \sim X}[f(x)]$ given samples $y_1, \dots, y_n$ we use%
$$\hat \psi = \frac{\sum_{i=1}^n w(y_i) f(y_i) }{\sum_{i=1}^n w(y_i)}$$
To generate approximate samples from $p(x)$, we resample from the set of samples in proportion to the importance weights.

If we iterate this procedure with a sequence of approximating distributions $q_1, \dots, q_k$, we get {\em Sequential Importance Sampling}.
If we resample at each stage, we get {\em Sequential Importance Resampling}.
If we additionally apply MCMC ``rejuvenation'' steps at each stage $i$ with a transition kernel that leaves the distribution $q_i$ invariant, we get {\em Sequential Monte Carlo}.

For Sequential Importance Sampling, the sum of the KL divergences between successive distributions controls the difficulty of sampling \citep{Freer:OudRha2O}.
If we can sample from the right coarse-grained distributions, we can reduce this difficulty, as illustrated in Figure \ref{fig:ctf-smc-illustration}.
With rejuvenation steps (SMC), the picture is more complex, but empirically, it is still the case that distributions that are closer together in KL generally make the sampling problem easier.
In particular, we expect that good coarse-to-fine sequences lead to better coverage of regions with high posterior probability, and that they enable more efficient pruning of low-probability regions.
A finite set of fine-grained particles may not cover the entire region, which can lead to a situation where all particles assign low probability to the next filtering step (particle decay).
A particle that has not been refined yet corresponds to distributions on fine-grained states, thus each such particle can cover a bigger region \citep{steinhardt2014filtering}.
Good coarse-to-fine sequences can allow us to prune entire parts of the state space in one go, only considering refinements of abstract states that have sufficiently high posterior probability \citep{kiddon2011coarse}.

\section{ALGORITHM}

Given a probabilistic program, our algorithm builds a coarse-to-fine program with the same marginal distribution as the original program, but with additional latent structure corresponding to coarsened versions of the program.

We will assume that the user provides a \lstinline{coarsenValue} function that describes how values map to more abstract values. Iterating this function leads to multiple levels of coarsened values. Our goal then is to construct a version of the original program that operates over values coarsened $N$ times. We will preserve the basic flow structure of the program, and thus we only need to specify how each primitive construct in the program is lifted to the space of coarsened values. The tricky part is to construct these lifted components such that the final marginal distribution is preserved. We use two ideas to accomplish this. First, we replace each unconditional elementary distribution at a given location with a distribution that depends on the coarser value of the same location, but such that the marginal over this coarser value yields the original distribution. Second, we treat lifted \lstinline{factor}s as only approximations useful for guiding inference, which are then canceled by an extra factor inserted at the next-finer level. With this scheme, only the finest-level factors contribute to the final score. This gains us flexibility over the lifting of primitive functions: lifted functions (that ultimately flow only to factor statements) only need to have similar behavior to their original; deviations will be corrected by the cancellation of factors.

In the next few subsections, we introduce heuristic factors, the inputs that the program transform requires, how the model syntax is transformed, and how each of the components of the lifted model works: constants, random variables, factors, and primitive functions.

\subsection{HEURISTIC FACTORS}
\label{sec:heuristic-factors}

A {\em heuristic factor} is a factor that is introduced for the purpose of guiding incremental inference algorithms such as particle filtering and best-first enumeration \citep{dippl}.
Its distinguishing characteristic is that an equivalent, canceling factor is inserted at a later position in the program in order to leave the program's distribution invariant.
In other words, the pair of statements \mbox{\lstinline{factor(s)}} and \mbox{\lstinline{factor(-s)}} together has no effect on the meaning of a model; its only effect is in controlling how inference algorithms explore the state space.

For example, Figure \ref{fig:heuristic-factors} shows a way to rewrite the program in Figure \ref{fig:probabilistic-program} in a way that initially assigns higher weight to program executions where $x$ is close to the true observation $7$. This is a heuristic, since---depending on the outcome of the coin flip---it may be $y$ which is observed, in which case there is no pressure for $x$ to be close to $7$.

The coarse-to-fine transform introduces heuristic factors that guide sampling on coarse levels towards high-probability regions of the state space without changing the program's distribution.

\subsection{PREREQUISITES}
\label{sec:prerequisites}

The main inputs to the transform are a model, given as code for a probabilistic program, and a pair of functions \lstinline{coarsenValue} and \lstinline{refineValue}.

The main constraint on the model is that all ERPs need to be independent, i.e., do not take parameters that depend on other ERPs. If the support of each ERP is known, this can be achieved using a simple transform that replaces each dependent ERP with a maximum-entropy ERP, and adds a dependent factor that corrects the score. That is, we transform
\begin{lstlisting}
  var x = sample(originalERP, params)
\end{lstlisting}
to
\begin{lstlisting}
  var x = sample(maxentERP)
  factor(originalERP.score(x, params) -
         maxentERP.score(x))
\end{lstlisting}

This transform leaves the model's distribution unchanged and greatly simplifies the coarsening of ERPs, but reduces the statistical efficiency of the model. This statistical inefficiency can potentially be addressed by merging \lstinline{sample} and \lstinline{factor} statements into \lstinline{sampleWithFactor} \citep{dippl} after the coarse-to-fine transform =.

The model is annotated with the name of the main model function (which defines the marginal distribution of interest) and a list of names of ERPs, constants, and functions (compound, primitive, score, and polymorphic; see below) to be lifted.

The main parameters that control the coarse distributions are the user-specified functions \lstinline{coarsenValue} and \lstinline{refineValue}. The function \lstinline{coarsenValue} maps a value to a coarser value; the function \lstinline{refineValue} maps a coarse value to a set of finer values.
To generate values on abstraction level $i$, we iterate the \lstinline{coarsenValue} function $i$ times.
We require that the two functions are inverses in the sense that
\mbox{$v \in $ \lstinline{refineValue}$(V)$} $\Leftrightarrow $ \mbox{\lstinline{coarsenValue}$(v) = V$} for all $v$ and $V$. 

If a model has multiple different types of variables for which inference is needed, it is easy to define a polymorphic coarsening function that implements different behaviors for different types of values. In cases where different variables have the same type but different meaning or scale, we recommend using wrapper types to control which coarsening is used.

This paper does not address the task of finding good value coarsening functions. Instead, we ask: if such a function is given, how can we use it to coarsen entire programs so that we produce a sequence of coarse-to-fine models that is useful for SMC?

\subsection{MODEL TRANSFORM}

\begin{figure}[t]
    \centering
    \begin{lstlisting}[
      basicstyle=\footnotesize\ttfamily\color{dimgray},
      identifierstyle=\color{dimgray},
      style=hilight]
@var liftedUniformERP=liftERP(uniformERP)@
@var liftedDistance=liftScorer(distance)@

var noisyObserve = function(obs){
  var score = -3*@lifted@Distance(obs, 7)
  @lifted@Factor(score)
}

var model = function(){
  @store.base = getStackAddress()@
  var x = @lifted@UniformERP()
  var y = @lifted@UniformERP()
  var observation = flip(0.5) ? x : y
  noisyObserve(observation)
  return [x, y]
}

@var coarseToFineModel = function(level){
  store.level = level
  var marginalValue = model()
  if (level === 0) {
    return marginalValue
  } else {
    return coarseToFineModel(level - 1)
  }
}@\end{lstlisting}
\caption{The coarse-to-fine model corresponding to the model in Figure \ref{fig:probabilistic-program}. This model has the same marginal distribution as \ref{fig:probabilistic-program} and \ref{fig:heuristic-factors}, but samples it using the hierarchical process shown in Figure \ref{fig:ctf-smc-illustration}.}
    \label{fig:ctf-model}
\end{figure}

The transform adds a wrapper \lstinline{coarseToFineModel} that calls the model once for each coarsening level, from coarse to fine, each time setting the (dynamically scoped) variable \lstinline{store.level} (in the following, $\level$) to the current level.
The transform also replaces all ERPs, factors, primitive functions, and score functions with lifted versions that act differently depending on $\level$.
The coarse models only affect the fine-grained models through the side-effect of storing the values of their random choices and the weights of their factors in \lstinline{store}, which is used by the finer-grained models to conditionally sample {\em their} random choices and compute {\em their} factor weights.

The syntactic transform itself proceeds as follow:

\begin{enumerate}
\item For each ERP, primitive and score function, insert the corresponding lifted definition before the model definition. For example:\\
\lstinline{var liftedPlus = liftPrimitive(plus)}
  \item Rename all ERPs, factors, primitive and score functions to their corresponding lifted names in model and compound functions. For example, replace \lstinline{plus} with \lstinline{liftedPlus}.
  \item Wrap all constants. For example, replace $c$ with \lstinline[mathescape]{liftConstant($c$)}.
  \item As the first statement in the model, store the current $\address$, which is needed to compute relative addresses of random choices and factors later on: \\
    \lstinline{store.base = getStackAddress()}
  \item Add a wrapper \lstinline{coarseToFineModel} that calls the model once for each coarsening level (see Figure \ref{fig:ctf-model}).
\end{enumerate}

We will now describe the mechanisms behind lifted constants, random variables, factors, and primitive functions.

\subsubsection{Lifting Constants}

To lift a constant, we simply repeatedly coarsen it to the current level:

\begin{figure}[h!]
\begin{minipage}{1.0\linewidth}
\hrule height 1pt \vskip 0.04in
\textbf{Algorithm 2: Lifting constants}
\vskip 0.04in \hrule height .5pt \vskip 0.04in
\begin{algorithmic}
  \small
  \Procedure{liftedConstant}{$c$}
  \For{$i$=0; $i < \level$; $i$++}
  \State $c$ = coarsenValue($c$)
  \EndFor
  \State \Return $c$
\EndProcedure
\end{algorithmic}%
\hrule height .5pt \vskip 0.04in
\end{minipage}
\end{figure}%

\subsubsection{Lifting Random Variables}

\begin{figure}[t]
\begin{minipage}{1.0\linewidth}
\hrule height 1pt \vskip 0.04in
\textbf{Algorithm 1: Lifting ERPs}
\vskip 0.04in \hrule height .5pt \vskip 0.04in
\begin{algorithmic}
  \small
  \Procedure{sampleLiftedERP}{$e_0, l$}
  \State $v_1$ = store[erpName($\address$, $l+1$)]
  \If{$v_1$ {\bf is} undefined}
  \If{$l$ {\bf is} $0$}
  \State \Return $e_0$.sample()
  \Else
  \State \Return coarsenValue(sampleLiftedERP($e_0, l-1$))
  \EndIf
  \Else
  \State $\vec v$ = refineValue($v_1$)
  \State $\vec p$ = $\vec v$.map($\lambda$(v)\{\Return getERPScore($e_0, v, l$)\})
  \State \Return sampleDiscrete($\vec v, \vec p$)
  \EndIf
\EndProcedure
\State
\Procedure{liftERP}{$e_0$}
  \State $e_1$ = makeERP($\lambda$()\{$\textsc{sampleLiftedERP}$($e_0, \level$)\})
  \State \Return $\lambda$()\{
  \State \ind $v$ = sample($e_1$)
  \State \ind store[erpName($\address$, $\level$)] = $v$
  \State \ind \Return $v$
  \State \}
  \EndProcedure
  \end{algorithmic}%
\hrule height .5pt \vskip 0.04in
\end{minipage}
\end{figure}%

We lift each random variable to a sequence of variables, from coarse to fine, such that (1) on the coarsest level, we unconditionally sample from the original distribution, coarsened $N$ times; (2) at each level $n < N$, we sample from the set of valid refinements of the value of the next-coarser variable; (3) marginalizing out all coarser variables, the distribution at the finest level recovers the distribution of the original uncoarsened variable.

Let $D_{0}$ be the domain of an ERP with distribution $p_{0}(x)$, and let $D_{n} = \mathrm{cv}^{n}(D_{0})$ be the set of values arrived at by repeatedly applying the \lstinline{coarsenValue} function (written $\mathrm{cv}$ for short). We would like to decompose the original distribution $p_{0}(x)$ into a sequence of conditional distributions $q(x_{n}|x_{n+1})$ for random variables $x_{n} \in D_{n}$.
If we take:
\[
q(x_n|x_{n+1}) \propto \sum_{x_0} p(x_0) \mathds{1}_{ \mathrm{cv}^{n}(x_0)=x_n  \land \mathrm{cv}(x_n) = x_{n+1} }
\]
and for the coarsest level, $N$,
\[
q(x_N) = \sum_{x_0} p(x_0) \mathds{1}_{\mathrm{cv}^{N}(x_0)=x_N}
\]
then it is clear that we preserve the marginal distribution on $x_{0}$. That is:
\[
p(x_0) = \sum_{x_{1},\dots,x_{N}} q(x_{0}|x_{1})\cdots q(x_{N-1}|x_{N})q(x_{N}).
\]

Algorithm 1 shows how we implement sampling from such a decomposed ERP at a given level.
Note that, to look up the existing value at the next-coarser level, we identify random variables on different levels based on relative stack addresses (via \lstinline{erpName}); this is critical for models such as grammars with an unbounded number of random choices.
The implementation for computing the score of lifted ERPs is analogous (although this score is not needed for pure particle filtering, so we omit the details).
Both are parameterized by a function \lstinline{getERPScore} that estimates the total probability of the equivalence class of values that map to a given coarse value. These ERP scores for coarse values can be estimated by sampling refinements, via user-specified scoring functions (as in Section \ref{sec:application-mrf}), or using exact computation (as in Section \ref{sec:application-fhmm}).

\subsubsection{Lifting Factors}

We treat the lifted counterpart to factors as heuristic factors: the score on the next-higher level is subtracted out on the current level. By canceling out these factors when inference proceeds to finer-grained levels, we ensure that the overall distribution of the model remains unchanged---ultimately, only the base-level factors count. This incremental scoring process formalizes the intuition of increasing attention to detail as we move down the abstraction ladder. Like random variables, we identify factors based on relative stack addresses.

\begin{figure}[h!]
\begin{minipage}{1.0\linewidth}
\hrule height 1pt \vskip 0.04in
\textbf{Algorithm 3: Lifting factors}
\vskip 0.04in \hrule height .5pt \vskip 0.04in
\begin{algorithmic}
  \small
  \Procedure{liftedFactor}{$s$}
  \State $s_1$ = store[factorName($\address$, $\level+1$)] $\lor$ 0
  \State factor($s - s_1$)
  \State store[factorName($\address$, $\level$)] = $s$
\EndProcedure
\end{algorithmic}%
\hrule height .5pt \vskip 0.04in
\end{minipage}
\end{figure}%

\subsubsection{Lifting Primitive Functions}

When a primitive function $f$ is applied to a base-level value, it is deterministic. Now we are interested in lifting primitive functions to operate on coarse values. However, each coarse value corresponds to a set of base-level values. For different elements in this set, $f$ may return different values. This suggests that lifted versions of $f$ may be stochastic.
We wish to preserve the marginal distribution of the entire program, but because we have required ERPs to be unconditional and treat coarse factors as canceling heuristics, we have some latitude in how to lift the primitive functions.

Algorithm 4 shows one approach to the computation of such coarsened primitives. This algorithm is parameterized by a function \lstinline{marginalize}, which may be implemented using exact computation, sampling, etc, and which may cache its computations.

\begin{figure}[h!]
\begin{minipage}{1.0\linewidth}
\hrule height 1pt \vskip 0.04in
\textbf{Algorithm 4: Lifting primitive functions}
\vskip 0.04in \hrule height .5pt \vskip 0.04in
\begin{algorithmic}
  \small
  \Procedure{liftPrimitive}{$f$}
  \State \Return $\lambda$($\vec x$)\{
  \State \ind $e$ = marginalize($\lambda$()\{\
  \State \ind \ind $\vec x_0$ = $\vec x$.map((uniformDraw $\circ$ refineValue)$^{\level}$)
  \State \ind \ind $v = f(\vec x_0)$
  \State \ind \ind \Return coarsenValue$^{\level}(v$)
  \State \ind \})
  \State \ind \Return sample($e$);
  \State \}
\EndProcedure
\end{algorithmic}%
\hrule height .5pt \vskip 0.04in
\end{minipage}
\end{figure}%

Scoring functions---those that directly compute a score to be consumed by \lstinline{factor}---are a special case of primitive functions. We know that they return a number, so instead of sampling from the return distribution, we can simply take the expectation. We find that this leads to more stable, and hence useful, heuristic factors.

Algebraic data type constructors are another special case. In many circumstances (including Section \ref{sec:application-fhmm}), they can be treated as transparent with respect to coarsening. For example, it is frequently useful to define
\lstinline[mathescape]{coarsenValue([$x_1$, $x_2$, $\dots$])} as equivalent to \lstinline[mathescape]{[coarsenValue($x_1$), coarsenValue($x_2$), $\dots$]}.
More generally, if a function can apply to coarsened objects directly, it can be marked as {\em polymorphic}. In that case, no lifting is necessary. For example, if we have a function that computes the mean of a list of numbers, and if our coarsening maps lists of numbers to shorter lists (without changing the type of the list elements), then we have the option to mark this function as polymorphic.

\section{EMPIRICAL EVALUATION}

We introduced two complementary ideas: (1) Coarse-to-fine SMC (which can be useful even for a single random variable with a big state space, given an appropriate coarsening function) and (2) constructing coarse-to-fine models from probabilistic programs (using a value coarsening function to lift each of the components of a probabilistic program to operate on coarsened values without changing the marginal distribution of the program).

Our experiments mirror this structure: In the first set of experiments, we evaluate the benefits of coarse-to-fine SMC on an Ising model and its depth-from-disparity variation. These models use a single matrix-valued random variable and a single global factor (the energy function).

In the second set of experiments---on a Factorial Hidden Markov Model---we make liberal use of probabilistic program constructs: the model is defined as a recursive stochastic function, and the state transition function is implemented using the higher-order function \lstinline{map}. This set of experiments fully exercises our program transform, including the identification of corresponding coarse and fine random variables using relative stack addresses.

All models are expressed as programs, and all experiments are implemented within the same framework. For each experiment, we show how the average importance weight---essentially a lower bound on the normalization constant \citep{grosse:2009:online}---behaves over time.

Note that these experiments are preliminary: They do not yet employ rejuvenation steps, and hence only test the quality of the sequence of coarse-to-fine distributions in the setting of Sequential Importance Resampling. While the use of a coarse-to-fine sequence of importance distributions results in better performance than baseline, we have only evaluated the performance compared to a relatively weak baseline (importance sampling for the MRF models, particle filtering for the Factorial HMM model), and expect that the results are not impressive on an absolute scale. Indeed, as described in Section \ref{sec:limitations}, we believe that some modifications and further developments are needed to make this approach useful in practice.

\begin{figure*}[t]
  \centering
  \begin{subfigure}[t]{.3\textwidth}
    \centering
    \includegraphics[scale=.5]{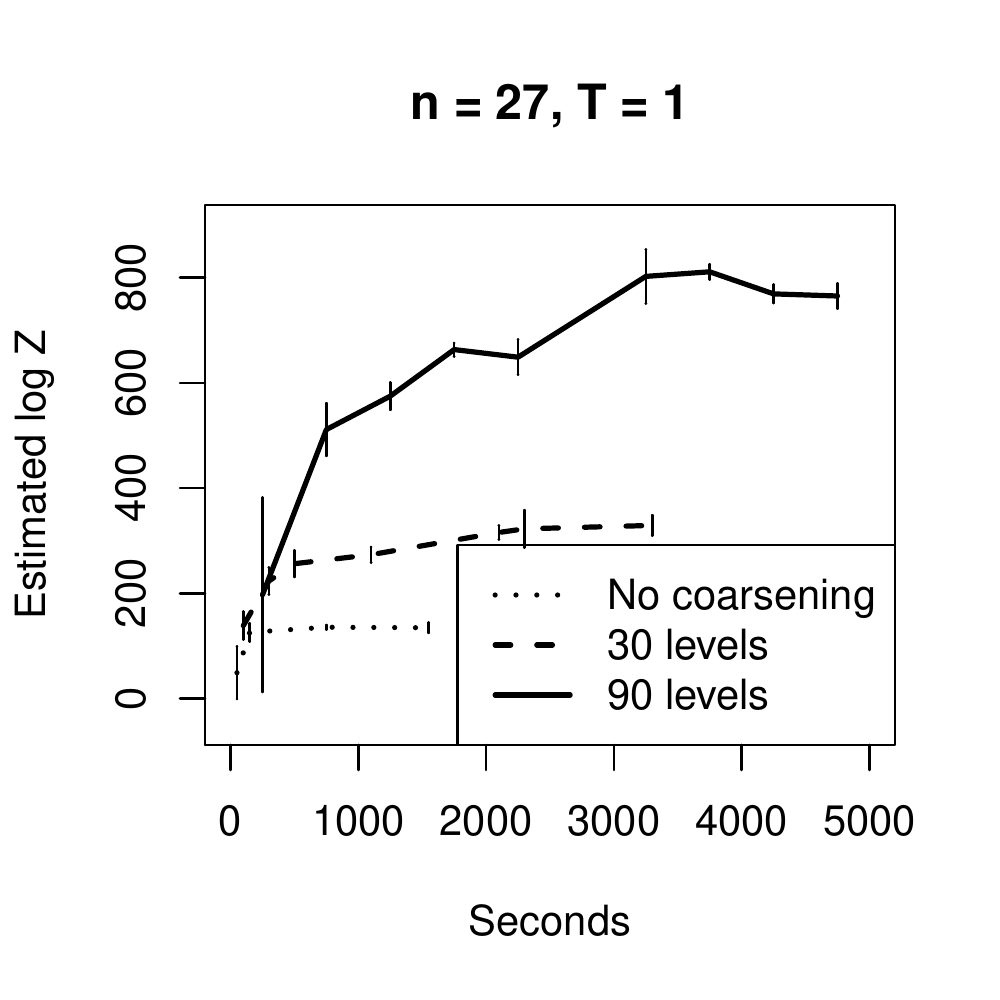}
    \caption{Ising at low temperature ($T = 1$)}
    \label{fig:isingT1}
  \end{subfigure}
  \hskip 1em
  \begin{subfigure}[t]{.3\textwidth}
    \centering
    \includegraphics[scale=.5]{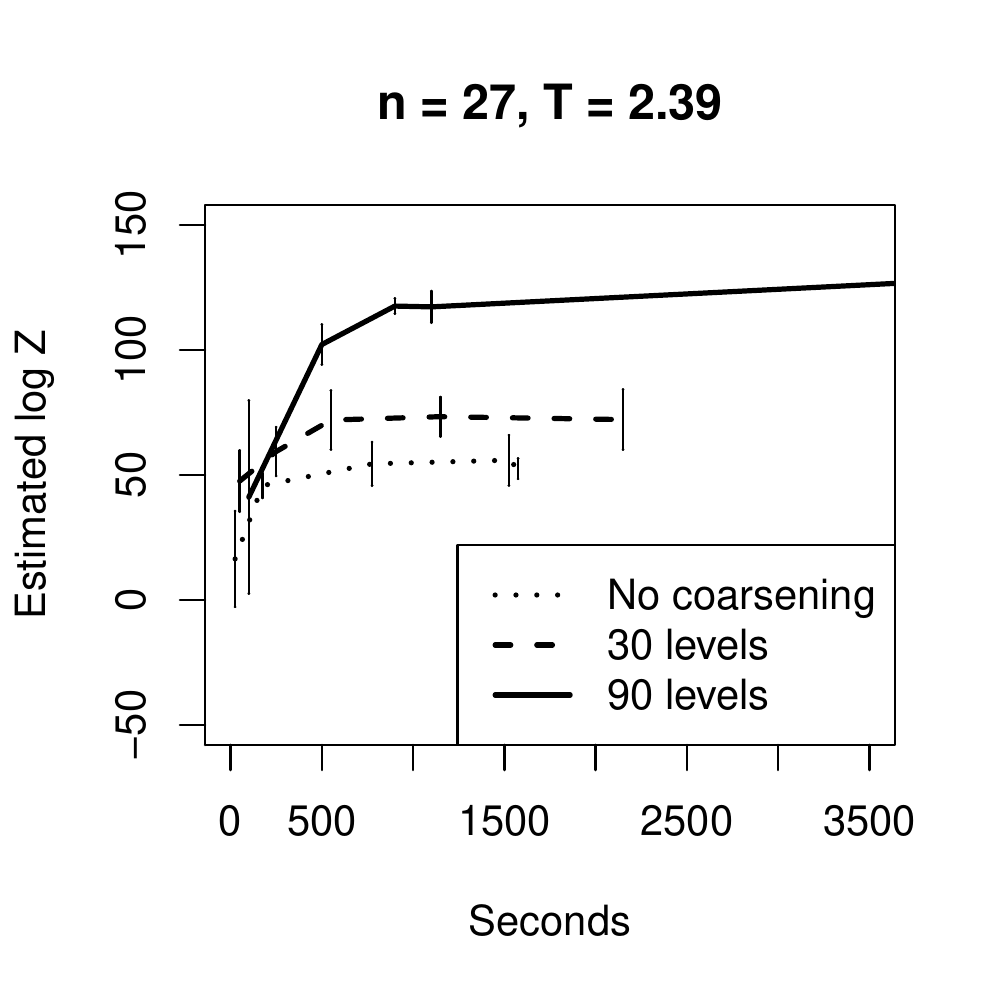}
    \caption{Ising at the critical state ($T = 2.39$)}
    \label{fig:isingTCritical}
  \end{subfigure}
  \hskip 1em
  \begin{subfigure}[t]{.3\textwidth}
    \centering
    \includegraphics[scale=.5]{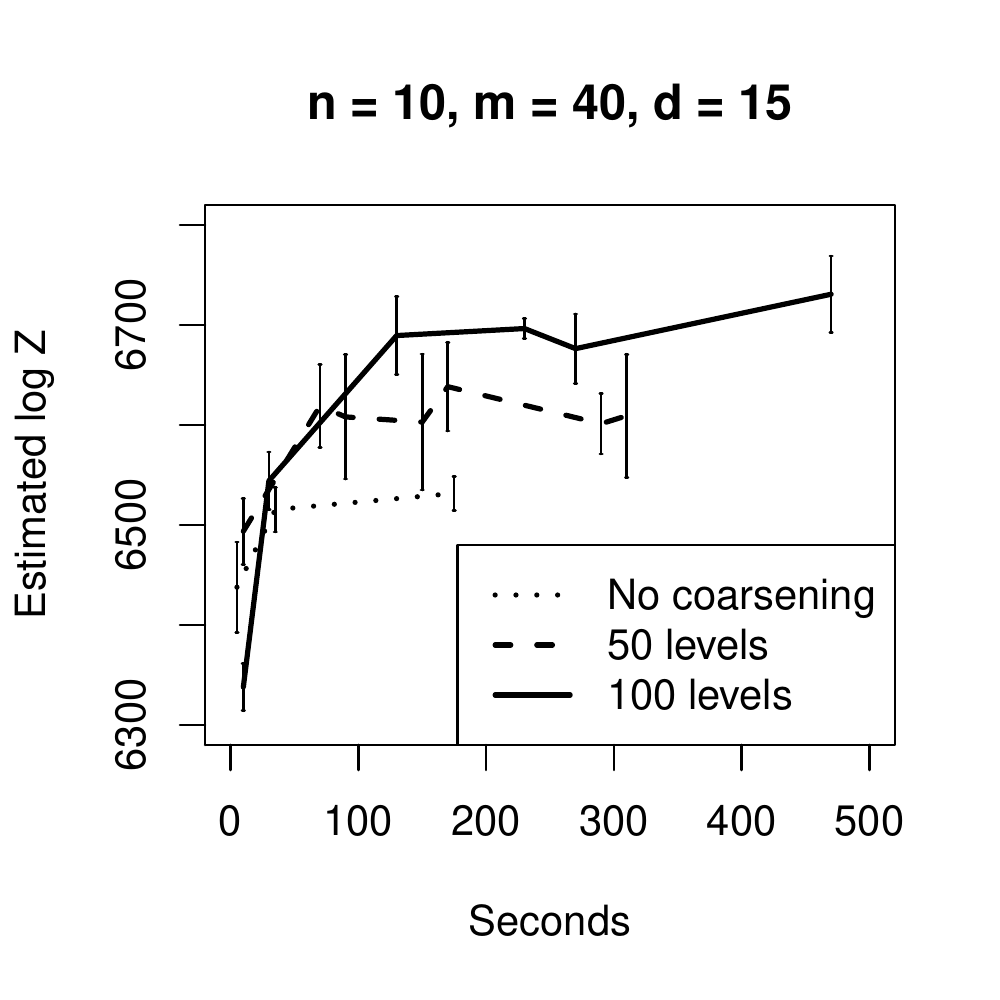}
    \caption{Depth-from-disparity}
    \label{fig:stereoResults}
  \end{subfigure}
  \caption{Quantitative inference results for Markov Random Field models}
\end{figure*}

\subsection{MARKOV RANDOM FIELDS}
\label{sec:application-mrf}

A number of applications in physics, biology, and computer vision can be modeled as Markov Random Fields (MRFs). These problems are unified in specifying a global energy function which, by virtue of the Markov property, depends only on the local neighborhoods of elements. Once this energy function is specified, it can be difficult to minimize; specialized optimization algorithms have been developed for particular domains \citep{szeliski2008comparative}, but there is no generally applicable solution.

The local neighborhood structure of the energy function, however, suggests that a coarse-to-fine transformation may be useful: if neighborhoods are coarsened into single representative values, then the energy can be minimized in this smaller space, using heuristic factors to guide search in the original space. We do not claim that our model transformation constitutes a solution in itself, but can be used in tandem with other algorithms to effectively reduce the search space. In this situation, we demonstrate the coarse-to-fine transformation on two simple MRFs: the Ising model and the stereo matching task.

\subsubsection{Ising model}

Coarse-to-fine transformations have a long history of applications in physics. When studying systems which interact across multiple orders of magnitude, such as fluids, ferromagnets, and metal alloys, it is intractable to work at the most fine-grained level. Since exact solutions do not exist, physicists developed a method called the \emph{renormalization group} \citep{wilson1975renormalization, wilson1979problems}, which effectively maps the fine-grained representation of a system onto a coarser but identically parameterized representation with similar properties.

One of the simplest testbeds for renormalization group methods is the 2-dimensional Ising model. The state space is an $n \times n$ lattice of cells, each of which can take one of two spin values, $\sigma_{i} \in \{-1, +1\}$. The \emph{energy} of a particular configuration of spins $\sigma$ is given by the Hamiltonian: $$\mathcal{H}(\sigma) = J \sum_{\langle i j\rangle} \sigma_i \sigma_j$$ where $J$ is the \emph{interaction constant} and $\langle i j \rangle$ indicates summing over all possible pairs of neighbors. Note that the number of possible configurations grows exponentially in $n$, rendering an exhaustive search for low-energy states impossible.

The interaction constant can be written as $J = 1/T$ where $T$ is the \emph{temperature} of the system. The configuration distribution takes different forms at different temperatures: we will conduct experiments at $T=1$, a low-temperature condition where spins prefer to globally align, and $T = 2.39$, the \emph{critical temperature}, where long-range correlations dominate. Above the critical temperature (e.g. for $T > 3$), cells become uncoupled and the energy distribution across configurations converges to uniform, so we focus on lower temperatures.

\begin{figure}[t]
  \centering
  \includegraphics[scale=.25]{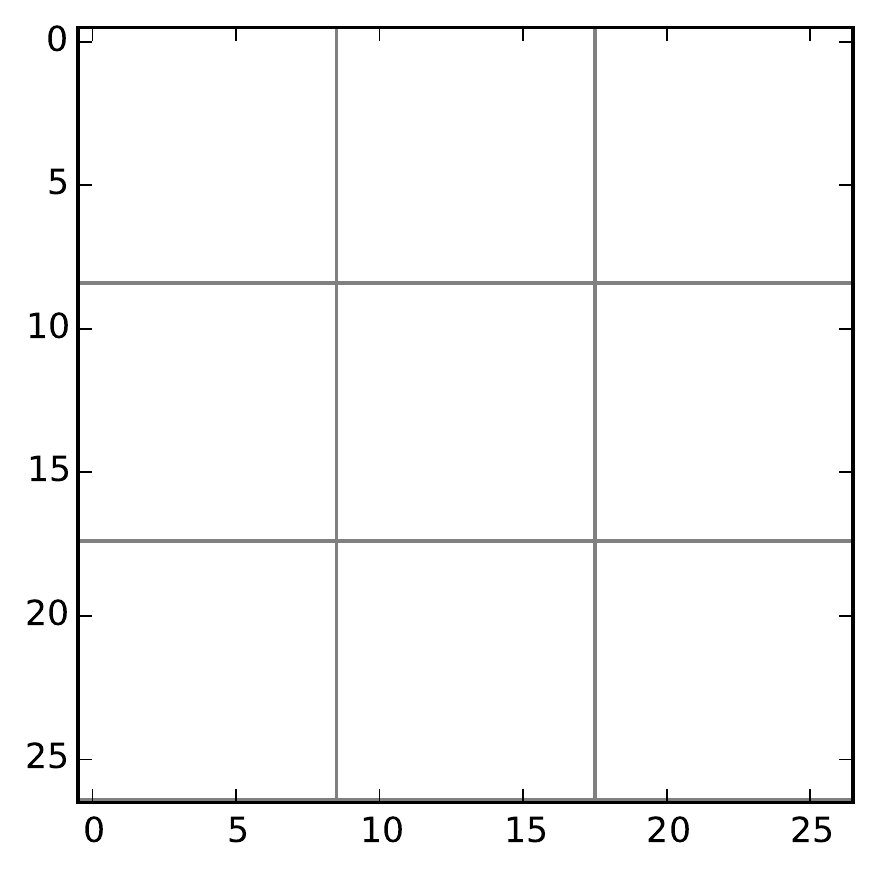}
  \hskip .5em
  \includegraphics[scale=.25]{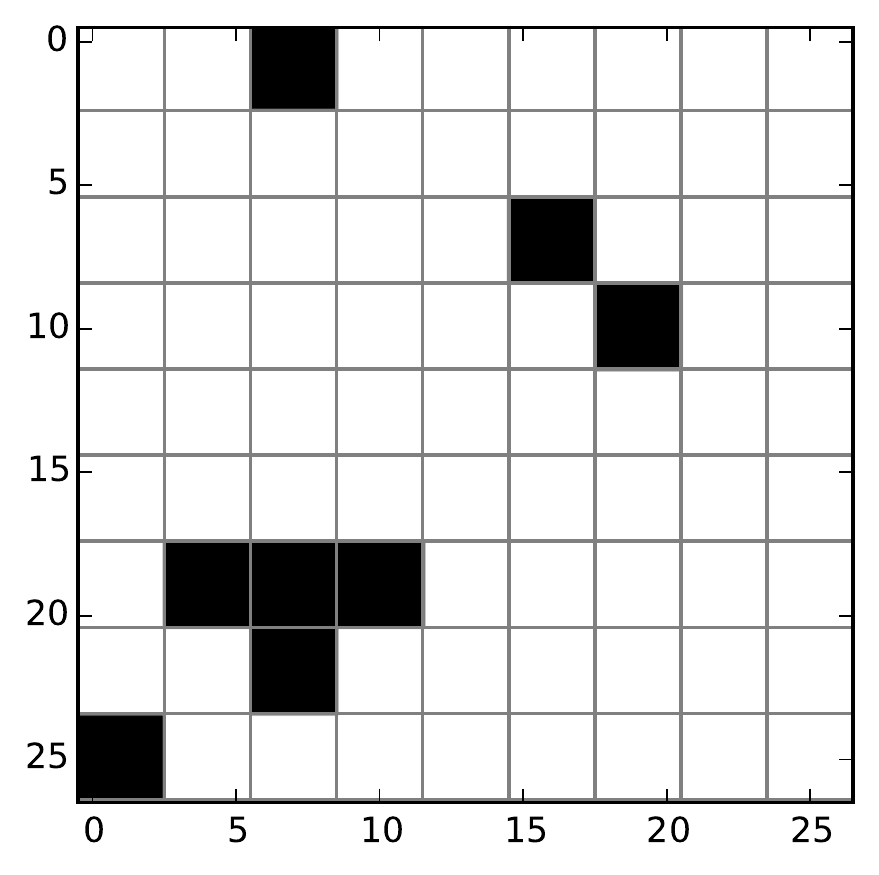}
  \hskip .5em
  \includegraphics[scale=.25]{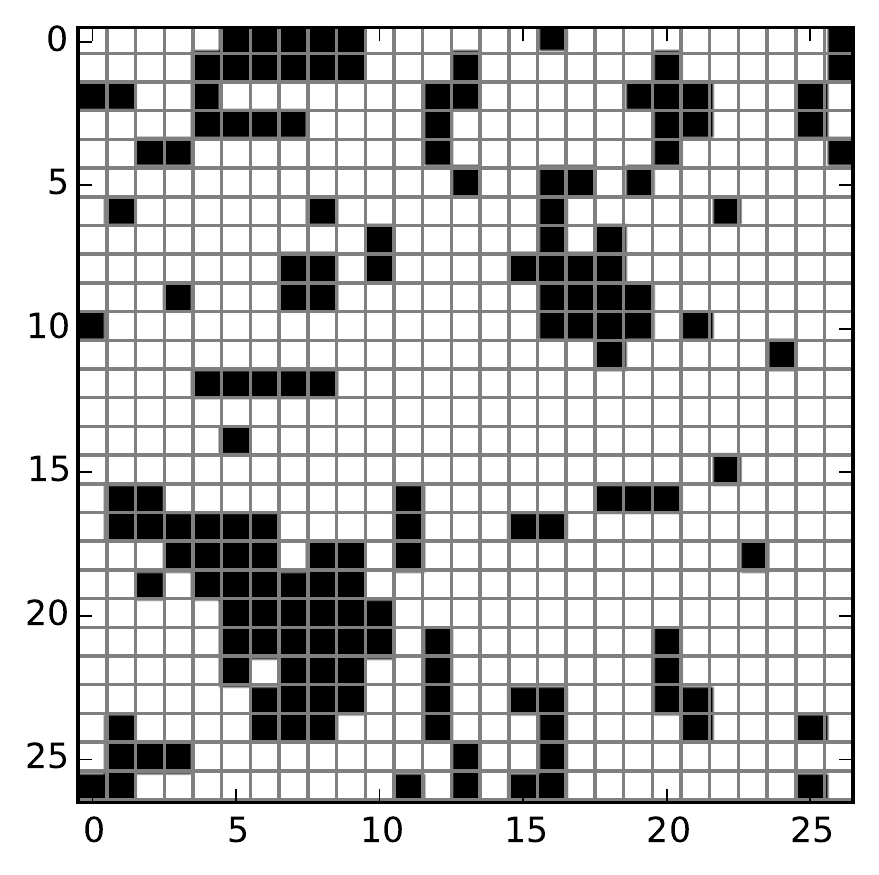}
  \caption{Coarsening at the critical state ($T = 2.39$)}
  \label{fig:coarseningLevels}
\end{figure}

We implemented the Ising energy-minimization problem as a simple probabilistic program, which first samples a set of spins and then factors based on the energy of that configuration. To apply our coarse-to-fine transformation, we used the spin-block majority-rule for our \texttt{coarsenValue} and \texttt{refineValue} functions. To coarsen, this rule replaced each $3 \times 3$ sub-lattice with its modal value (see Figure \ref{fig:coarseningLevels}). To refine a single cell in the coarse matrix, we considered the space of all 256 possible $3\times 3$ matrices that could coarsen to that value.
Note that our sequential refinement -- making many small choices instead of one big one -- differs from the typical renormalization group approach, which simultaneously replaces all sublattices and reweights the interaction constant $J$ accordingly.

For example, consider a fine-grained value such as\footnote{We use spins $\{1, 0\}$ here instead of $\{1, -1\}$ to make the matrices more readable.}:

\[
A = \begin{bmatrix}
0 & 1 & 0 & 0 & 0 & 0 \\
1 & 1 & 0 & 1 & 0 & 0 \\
1 & 1 & 1 & 1 & 0 & 0 \\
0 & 1 & 1 & 0 & 0 & 0 \\
0 & 0 & 0 & 0 & 0 & 1 \\
0 & 1 & 1 & 0 & 1 & 1
\end{bmatrix}
\]

Coarser values are partially-coarsened matrices, which can be represented as pairs of matrices, e.g.

\[
\texttt{cv}(A) = \left (
\begin{bmatrix}
* & * & * & 0 & 0 & 0 \\
* & * & * & 1 & 0 & 0 \\
* & * & * & 1 & 0 & 0 \\
0 & 1 & 1 & 0 & 0 & 0 \\
0 & 0 & 0 & 0 & 0 & 1 \\
0 & 1 & 1 & 0 & 1 & 1
\end{bmatrix},
\begin{bmatrix}
  1 & * \\
  * & *
\end{bmatrix}
\right )
\]

where each entry in the second matrix corresponds to a 3x3 block in the first matrix. In each coarsening step, we only coarsen a single such block. As a consequence, there are 90 (= 81 + 9) steps when we incrementally coarsen a 27x27 matrix first to 9x9, then to 3x3.

To facilitate this sequential refinement, we implemented a polymorphic energy function, which can directly return a score for such partially coarsened matrices without needing to refine all the way down to the most fine-grained level. When this energy function is applied to a matrix that contains cells at different levels of abstraction, the energy computation counts a single coarse cell as a neighbor for each surrounding finer cell (and vice versa).

We ran two experiments on $27 \times 27$ lattices to demonstrate the performance of our coarsened program at different levels of coarsening. In our first experiment, we set the temperature to $T=1$ and average 10 runs for both coarse-to-fine filtering (with $90$ and $30$ levels) and flat importance sampling. The $90$ level coarsening condition fully reduces the $27 \times 27$ lattice to a $3 \times 3$ lattice, and the $30$ level condition yields a partially coarsened matrix. In our second experiment, we set the temperature to $T = 2.39$ and run the same set of conditions. Figures \ref{fig:isingT1} and \ref{fig:isingTCritical} show the average importance weight for different levels of coarsening. We see that even intermediate levels of coarsening perform better than no coarsening, and that a full coarsening performs dramatically better than the other conditions. The best solution from the second experiment is shown in the right-most panel of Figure \ref{fig:coarseningLevels}, along with the two coarser configurations from which it was refined. This displays the characteristic local structure of the Ising model at critical temperatures.

\subsubsection{Stereo matching}

Another common application of MRFs is the stereo matching task \citep{scharstein2002taxonomy, BoykovVekslerZabih01_GraphCutsEnergyMinimization}. The goal is to estimate the disparity between two images, $\mathcal{I}$ and $\mathcal{I'}$, captured from slightly shifted viewpoints. This disparity map can be used to recover a rough measure of depth. As in the Ising case, we implement this task as a probabilistic program by sampling a lattice of disparity values, and factoring on its energy.

The energy function for a particular set of disparity values has two parts: (1) a \emph{smoothing} term penalizing distance between the values of neighbors and (2) a \emph{data cost} term penalizing each particular disparity value for discrepancies with the true data (as measured by comparing the difference in pixel intensities at the given discrepancy):

$$\mathcal{H}(d) = \sum_{\{p, q\} \in \mathcal{N}} V_{p,q}(d_p, d_q) + \sum_p C(p, d_p)$$

We denote the intensity of pixel $p$ in image $\mathcal{I}$ by $\mathcal{I}_p$. Since corresponding pixels should have similar intensities, we set our data cost term $C(p, d_p)$ as suggested by \cite{BoykovVekslerZabih01_GraphCutsEnergyMinimization}, taking the absolute difference between $\mathcal{I}_p$ and $\mathcal{I}_{p+d_p}'$. To reduce sensitivity to variability in image sampling, we interpolate between neighboring intensities in the neighborhood $x \in (d-0.5, d+0.5)$ and take the minimum. For our smoothing function, we use the truncated squared error: $$V_{p,q} (d_p, d_q) = \min((d_p - d_q)^2, V_{\max})$$ with $V_{\max} = 5$.

We implemented energy minimization for the stereo matching model analogous to the Ising model, but with different coarsening and refinement functions: coarsening replaces a $2 \times 2$ sublattice with its mean and standard deviation; refinement returns the set of all possible $2 \times 2$ lattices with the given mean and standard deviation that could have occurred at the given abstraction level.

Figure \ref{fig:stereoResults} shows that in a comparison between coarse-to-fine SMC and importance sampling, SMC finds lower-energy states more efficiently for a 10x40 cropped pair of images from the Middlebury dataset \citep{scharstein2002taxonomy}, although we expect that the absolute quality of the states found is not very good for either algorithm.

\begin{figure*}[t]
  \vskip -3em      
  \centering
  \begin{subfigure}[t]{.45\textwidth}
    \centering
    \includegraphics[scale=.7]{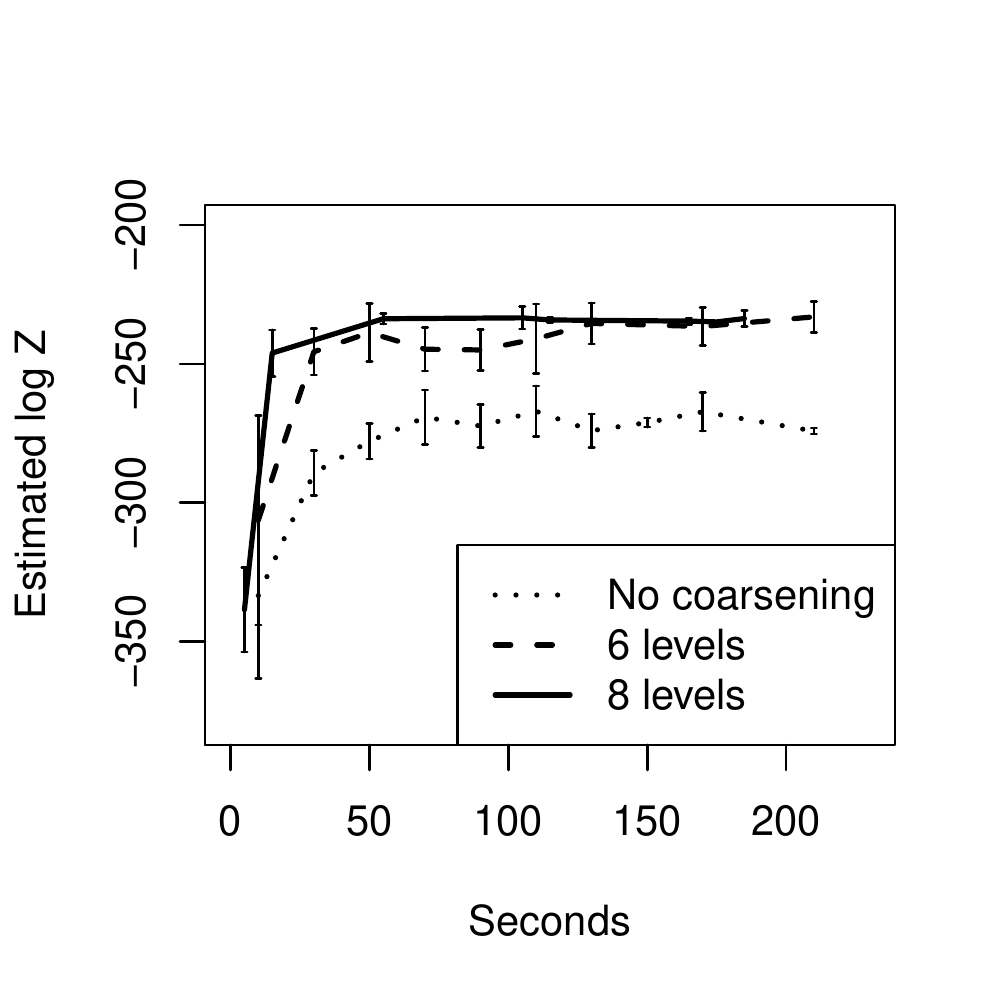}
    \caption{SMC using a coarse-to-fine model finds more probable samples earlier on than SMC without coarsening.}
    \label{fig:factorial-hmm-time}
  \end{subfigure}
  \hskip 2em
  \begin{subfigure}[t]{.45\textwidth}
    \centering
    \includegraphics[scale=.7]{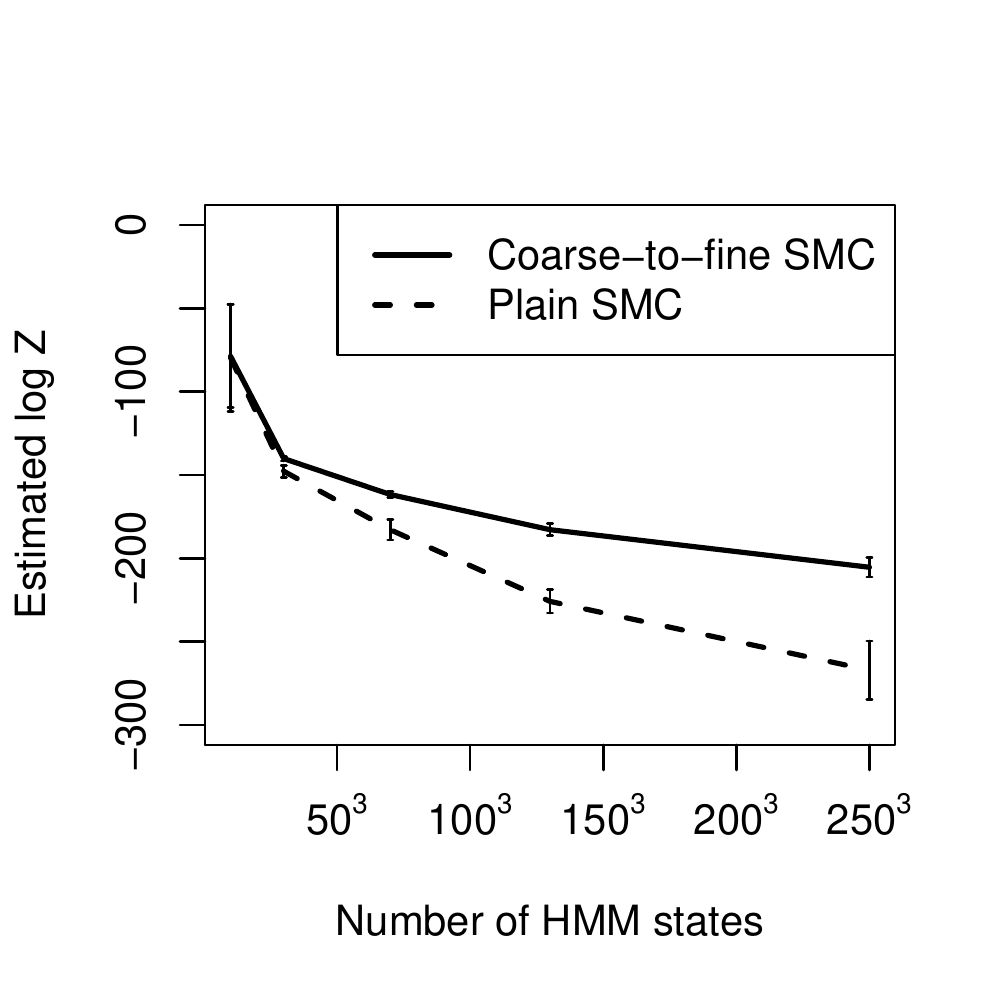}
    \caption{For small numbers of states, coarse-to-fine is indistinguishable from plain particle filtering. As the number of states grows, coarse-to-fine is able to provide better solutions in the same amount of time.}
    \label{fig:factorial-hmm-statespace}
  \end{subfigure}
  \caption{Inference results for a Factorial HMM}
\end{figure*}

\subsection{FACTORIAL HMM}
\label{sec:application-fhmm}

In our second example, we test the hypothesis that abstractions are useful as a means to avoid particle collapse in large state spaces. For this reason, we chose the Factorial HMM, a model with a large effective state size even within a single particle filter step. The Factorial HMM is a HMM where the state factors into multiple variables \citep{Ghahramani:1997id}. If there are $M$ possible values for each latent state, and $k$ state variables per time step, then the effective state size is $M^k$. If only few of these have high probability, then even for moderate $M$ and $k$ it is possible that there are not sufficiently many fine-grained particles to cover all regions of high posterior probability.

In our first experiment, we use a Factorial HMM with $3$ variables per step, $256$ possible state values per variable, and $6$ observed time steps. We run both coarse-to-fine filtering (with $6$ and $8$ abstraction levels) and flat filtering and average $10$ runs.

We coarsen the Factorial HMM by merging some state and observation symbols. To test the hypothesis that coarse-to-fine inference will work best when abstractions match the dynamics of the model, we generate transition and observation matrices with approximately hierarchical structure as follows. Enumerate state $1$ to $N$. 
For states $i$ and $j$, we let the transition probability be approximately proportional to $2^{-|i-j|}$. Similarly, for state $i$, the probability of generating observation $k$ is proportional to $2^{-|i-k|}$.

Each state consists of three substates, and each substate is chosen from $\{1,2,\dots,256\}$. Coarsening maps numbers to successively greater intervals. For example, coarsening maps $(5, 11, 56)$ to $([5, 6], [11, 12], [55, 56])$, and on the next-coarser level to $([5, 8], [11, 14], [55, 58])$. Figure \ref{fig:factorial-hmm-time} shows that plain particle filtering consistently underestimates the true normalization constant relative to coarse-to-fine filtering.

In our second experiment (Figure \ref{fig:factorial-hmm-statespace}), we compare the behavior of flat and coarse-to-fine filtering as the number of HMM states increases from $2^3$ to $256^3$. As before, we keep the runtime constant.
For small numbers of states, plain and coarse-to-fine SMC give very similar estimates. As the number of states grows, the difference between coarse-to-fine and plain filtering grows as well, indicating that coarse-to-fine is most useful in large state spaces.

\section{DISCUSSION}

\subsection{WHEN DOES COARSE-TO-FINE INFERENCE HELP?}

It is generally difficult to compute or estimate the posterior probability of a set of (program) states.
However, this is precisely what is required for coarse-to-fine inference to work: when we evaluate a program on a coarse level, we need to estimate for each coarse value how likely its refinements are under the posterior.
This suggests that settings where coarse-to-fine inference works have special characteristics that make such estimation feasible.
We now name a few.

First, the given program may satisfy independence assumptions that make estimating posterior probabilities feasible. For example, for the program shown in Figures \ref{fig:ctf-smc-illustration} and \ref{fig:probabilistic-program}, the score function only depends on one of $x$ and $y$ at a time; hence, we can independently compute the estimated score for the refinements of $x$ and $y$, and use this information in computing the estimated scores for abstract values of both.

Second, we may be in a setting where the type of a coarse value matches the type of its refinements. In that situation, ``polymorphic'' score- and primitive functions may be a cheap heuristic for estimating the posterior probability of a coarse state. For the Ising model, the energy function satisfies this criterion up to parameterization.

Third, coarse-to-fine may be particularly useful in the amortized setting \citep{Stuhlmuller:2013ab}. Learning the conditional distributions associated with lifted primitive functions is one instantiation of ``learning to do inference''. This is particularly feasible for smooth state spaces, where one can effectively estimate entire distributions from a few samples.

Another answer to the question of when coarse-to-fine helps is to point out that this depends on what inference algorithm is used. For inference by enumeration, exact coarsening (i.e., coarsening within values that have the {\em same} posterior probability) is useful for increasing computational efficiency. By contrast, for sequential Monte Carlo methods, it is frequently more desirable to merge states with {\em different} posterior probability, as this smoothes the state space and thus increases statistical efficiency.

\subsection{CURRENT LIMITATIONS}
\label{sec:limitations}

The system as presented has three main limitations. We will now describe these limitations and outline how they can be addressed.

First, as discussed in Section \ref{sec:prerequisites}, the coarse-to-fine transform can only be applied to models where all ERPs are independent. We have described how to transform models into this form and referred to the technique of merging \lstinline{sample} and \lstinline{factor} statements described in \cite{dippl} as a tool for recovering statistical efficiency lost in such a transform. We have not yet employed this merging in our experiments, which makes the results more difficult to interpret. We expect that it would be feasible to develop a version of the coarse-to-fine transform that directly operates on dependent ERPs.

Second, single-site MCMC rejuvenation steps are of very limited use in the current setup. This is due to the combination of (a) computing coarsened values by binning fine-grained values (using the user-provided \lstinline[mathescape]{coarsenValue} function) and (b) using this coarsening to build a hierarchical model to which standard SMC algorithms can be applied. In this hierarchical model, each fine-grained value $v$ only has non-zero probability if the corresponding coarse value is set to \lstinline[mathescape]{coarsenValue($v$)}. As a consequence, we reject all MCMC steps that change this coarse-grained value to anything else, and likewise all steps that change $v$ to $v'$ such that \lstinline[mathescape]{coarsenValue($v$) $\neq$ coarsenValue($v'$)}. These strong dependencies between levels of abstraction can be avoided if we only use each level in the sequence of coarse-to-fine models as an importance sampler for the next level instead of constructing a coarse-to-fine model.

Third, coarsening only applies to individual variables and values, not to sets of variables. Consider the Ising model: while we implemented it using a matrix-valued random variable, a more natural implementation would use independent Bernoulli variables for each of the matrix entries. Our current scheme cannot express coarsenings that apply to such sets of variables. While it is easy to see how our scheme can be extended to coarsening multiple variables within the same control-flow block (by transforming multiple variables to a single tuple-valued variable), generic coarsening of variables that are created in different control flow blocks will require new innovations.

\subsection{RELATED AND FUTURE WORK}

The work in this paper is related to and inspired by a broad background of work on coarse-to-fine and lifted inference, including (but not limited to) work by \citet{Charniak:2006ut} on coarse-to-fine inference in PCFGs, work on coarse-to-fine inference for first-order probabilistic models by \citet{kiddon2011coarse}, and attempts to ``fatten'' particles for filtering with broader coverage \citep{steinhardt2014filtering, Kulkarni2014}. Outside of machine learning, we take inspiration from {\em Approximate Bayesian Computation} (conditioning on summary statistics) and {\em renormalization group} approaches to inference in Ising models \citep{Lyman2006}.

This approach opens up many exciting research directions, including (1) understanding the relation to abstract interpretation, and Galois Connections specifically \citep[e.g.,][]{monniaux2000abstract,cousot2012probabilistic}, (2) automatically deriving coarsenings for hierarchical Bayesian models, (3) learning good coarsenings, and efficient learning of approximations for coarsened primitive and score functions, and (4) coarsening (merging) multiple variables across blocks, potentially via flow analysis.

We expect that the most interesting applications of coarse-to-fine approaches to efficient inference are yet to come.

\subsubsection*{Acknowledgements}

This material is based on research sponsored by DARPA under agreement number FA8750-14-2-0009. The U.S. Government is authorized to reproduce and distribute reprints for Governmental purposes notwithstanding any copyright notation thereon.

RXDH is supported by a NSF Graduate Research Fellowship and a Stanford Graduate Fellowship.

\subsubsection*{References}

\bibliography{ctf-arxiv2015}
\bibliographystyle{plainnat}

\end{document}